\documentclass[11pt, a4paper, logo, onecolumn, copyright, nonumbering]{deepmind}

\usepackage[authoryear, sort&compress, round]{natbib}
\usepackage{amsmath,amsfonts,bm}

\newcommand*{\defeq}{\stackrel{\text{def}}{=}}

\def\eqref#1{equation~\ref{#1}}

\def\1{\bm{1}}

\newcommand{\transitions}{T}

\def\vx{{\bm{x}}}
\def\vy{{\bm{y}}}

\DeclareMathAlphabet{\mathsfit}{\encodingdefault}{\sfdefault}{m}{sl}
\SetMathAlphabet{\mathsfit}{bold}{\encodingdefault}{\sfdefault}{bx}{n}

\def\gA{{\mathcal{A}}}

\def\gL{{\mathcal{L}}}
\def\gM{{\mathcal{M}}}

\def\gS{{\mathcal{S}}}

\newcommand{\calD}{{\cal D}}

\newcommand{\data}{\calD}

\newcommand{\E}{\mathbb{E}}

\usepackage{floatrow}
\usepackage{varwidth}
\usepackage[]{algorithm2e}
\usepackage{algpseudocode}
\usepackage{bbm}
\usepackage{wrapfig}
\usepackage{siunitx}

\title{Reinforced Self-Training (\rest) for Language Modeling}%

\correspondingauthor{ca9lar@gmail.com}

\keywords{Offline RL, reinforcement learning, RL from human feedback, language, natural language processing, machine translation}

\newcommand{\rest}{\emph{ReST}}
\newcommand{\grow}{\texttt{Grow}}
\newcommand{\improve}{\texttt{Improve}}
\newcommand{\BC}{\texttt{BC}}

\definecolor{dm-blue-500}{RGB}{0, 69, 177}
\definecolor{dm-purple-500}{RGB}{105,50,230}

\reportnumber{001} %

\author{
{\Authfont
Caglar Gulcehre\textsuperscript{*\textdagger,1},
Tom Le Paine\textsuperscript{*\textdagger,1},
Srivatsan Srinivasan\textsuperscript{*\textdagger,1},
Ksenia Konyushkova\textsuperscript{\textdagger,1}, 
Lotte Weerts\textsuperscript{\textdagger,1}}
\leavevmode \protect\\[1em]
{\Authfont
Abhishek Sharma\textsuperscript{\textdagger,1}, 
Aditya Siddhant\textsuperscript{\textdagger,1}, 
Alex Ahern\textsuperscript{1}, 
Miaosen Wang\textsuperscript{1},
Chenjie Gu\textsuperscript{1}}, 
\leavevmode \protect\\[1em]
{\Authfont
Wolfgang Macherey\textsuperscript{2},
Arnaud Doucet\textsuperscript{1},
Orhan Firat\textsuperscript{\textdagger,1}, 
Nando de Freitas\textsuperscript{1}}
\leavevmode \protect\\[1em]
{\Affilfont
\textsuperscript{*}Contributed equally, \textsuperscript{\textdagger}Core contributors\protect\\
\textsuperscript{1}Google DeepMind, \textsuperscript{2}Google Research 
}}

\begin{abstract}

Reinforcement learning from human feedback (RLHF) can improve the quality of large language model's (LLM) outputs by aligning them with human preferences. We propose a simple algorithm for aligning LLMs with human preferences inspired by growing batch reinforcement learning (RL), which we call \textit{Reinforced Self-Training} (\rest{}). Given an initial LLM policy, \rest{} produces a dataset by generating samples from the policy, which are then used to improve the LLM policy using offline RL algorithms. \rest{} is more efficient than typical online RLHF methods because the training dataset is produced offline, which allows data reuse. While \rest{} is a general approach applicable to all generative learning settings, we focus on its application to machine translation. Our results show that \rest{} can substantially improve translation quality, as measured by automated metrics and human evaluation on machine translation benchmarks in a compute and sample-efficient manner.

\end{abstract}

\begin{document}
\maketitle
\section{Introduction}

\begin{wrapfigure}[19]{r}{0.4\textwidth}
    \begin{center}
    \vspace{-12mm}
    \includegraphics[width=0.8\linewidth]{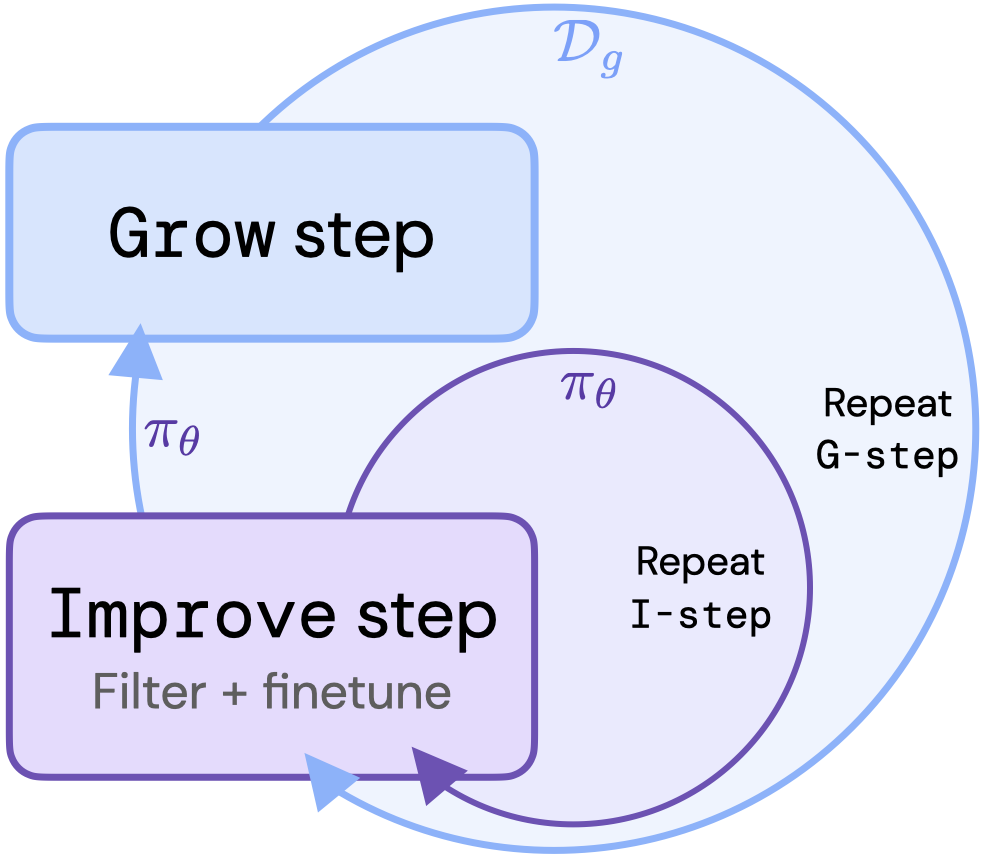}
    \end{center}
    \caption{{\textbf{\rest{} method.}} During \grow{} step, a policy generates a dataset. At \improve{} step, the filtered dataset is used to fine-tune the policy. Both steps are repeated, \improve{} step is repeated more frequently to amortise the dataset creation cost.}
    \label{fig:first_figure}
\end{wrapfigure}

Large language models (LLMs) have demonstrated impressive abilities in generating high-quality text and in solving numerous language tasks \citep{bubeck2023sparks,brown2020language,rae2021scaling}. These models are trained to maximize the likelihood of the next token autoregressively using massive amounts of text and compute \citep{srivastava2022beyond,hoffmannempirical}. However, \cite{perez2022red} showed that producing text with high likelihood does not necessarily align well with human preferences on various tasks. Without proper alignment, the language models can also output unsafe contents with harmful consequences. Moreover, aligning LLMs helps to improve on other downstream tasks \citep{ouyang2022training}. \emph{Reinforcement learning from human feedback} (RLHF) aims to address the alignment problem by using human preferences \citep{glaese2022improving,wu2021recursively,stiennon2020learning}. Typically, human feedback is used to learn a reward model, which is then used to fine-tune LLM with a reinforcement learning (RL) objective.

RLHF methods often rely on online RL methods such as PPO \citep{schulman2017proximal} and A2C \citep{mnih2016a2c}. Online training requires  sampling from the updated policy and scoring the samples with the reward model many times during training. The computational cost of dealing with a continual flow of new samples becomes a limitation of online methods, especially when the sizes of the policy and reward networks grow. Moreover, these methods are prone to reward ``hacking'' \citep{skalse2022rewhack}, and prior works \citep{glaese2022improving} explored model regularization to mitigate this issue. Alternatively, offline RL methods learn from a fixed dataset of examples and, thus they are more computationally efficient and less prone to reward hacking. However, the quality of the policy learnt offline inevitably depends on the properties of the offline dataset \citep{gulcehre2021regularized,fu2020d4rl}. As a result, carefully curated datasets become very important for the success of offline RL. Otherwise, the performance gains over supervised learning may be limited \citep{kumar2021should}. Concurrent to our work, \citep{rafailov2023direct} proposed a method called DPO (Direct Preference Optimization) that can make use of offline data to align an LM with human preferences.

We frame the alignment problem of a language model as a growing batch RL problem \citep{lange2012batch}. Specifically, our Reinforced Self-Training (\rest) method includes two loops: in the inner loop (\improve{}), we improve the policy on a fixed dataset and in the outer loop (\grow{}), we grow the dataset by sampling from the latest policy (see Figure~\ref{fig:first_figure}). In this work we consider conditional language modelling, then the steps of \rest{} are as follows: 
\vspace{-3mm}
\begin{enumerate}
	\item \grow{} (\texttt{G}): The language model policy (initially, a supervised policy) is used to generate multiple output predictions for each context to augment the training dataset. 
	\item \improve{} (\texttt{I}): We rank and filter the augmented dataset with a scoring function. We use a learned reward model trained on human preferences as the scoring function in our experiments. Then, the language model is fine-tuned on the filtered dataset with an offline RL objective. This step can be repeated with an increasing filtering threshold. The final policy is then used in the next \grow{} step.
\end{enumerate}
\vspace{-3mm}

\rest{} is a general approach that allows different offline RL losses to be used in the inner loop when executing the \improve{} steps. To put it in practice, one only needs the ability to: i) sample from a model efficiently, ii) score the model's samples. \rest{} provides several advantages over typical RLHF methods with online or offline RL:
\vspace{-3mm}
\begin{itemize}
    \item The computational burden is significantly reduced compared to online RL thanks to the output of \grow{} step being exploited across several \improve{} steps.
    \item The quality of the policy is not restricted by the quality of the original dataset (as in offline RL) since new training data is sampled from an improved policy during the \grow{} step.
    \item It is easy to inspect the data quality and potentially diagnose alignment issues, \emph{e.g.,} reward hacking, as the \grow{} and \improve{} steps are decoupled.
    \item The approach is simple, stable and has only a small number of hyperparameters to tune.
\end{itemize}
\vspace{-2mm}

We explain the details of our proposed \rest{} approach in Section \ref{sec:rest}. Then, we present our experimental results on machine translation benchmarks in Section \ref{sec:experiments}. 
Machine translation is a sequence-to-sequence learning problem \citep{sutskever2014sequence}, which is usually formulated as conditional language modelling where the context for conditioning is a sentence in a foreign language (source). We chose machine translation because i) it is an impactful application with strong baselines and a well-defined evaluation procedure, ii) several existing reliable scoring and evaluation methods are available for the use as a reward model \citep{freitag-etal-2022-results}. In our experiments, we compare several offline RL algorithms on the IWSLT 2014 \citep{cettolo2014report} and WMT 2020 benchmarks \citep{koehn2020findings} as well as more competitive, high-fidelity internal benchmarks on Web Domain. In our experiments \rest{} significantly improves reward model scores on test and validation sets. Furthermore, according to human raters, \rest{} generates higher quality translations compared to a supervised learning baseline.

\section{Preliminaries}
\label{sec:preliminaries}
A conditional language model produces an output sequence $\vy = \left(y_1, y_2, ....y_T\right)$ given a context (or source input) $\vx = \left(x_1, x_2, ...x_L\right)$, where the tokens $x_l, y_t$ belong to a chosen vocabulary. A language generation policy $\pi$ in an auto-regressive model characterized by a conditional probability distribution parameterized by $\theta$ as 
\[
\pi_{\theta}(\vy\mid \vx)=\prod_{t=1}^T \pi_{\theta}(y_t\mid \vy_{1:t-1}, \vx),
\]
with the convention $\vy_{1:0}=\emptyset$ and $\vy_{1:t-1} = \left(y_1, y_2, ....y_{t-1}\right)$.

Let $p(\vx,\vy)=p(\vx)p(\vy|\vx)$ denote the data distribution. A given dataset $\data$ consists of samples from this distribution:
\[
\data = \{ \; (\vx^i,\vy^i)|_{i=1}^{N} \;\; \mbox{such that} \;\; \vx^i \sim p(\vx), \; \vy^i \sim p(\vy|\vx=\vx^i) \; \}. 
\]
Given this dataset, the supervised policy is trained by minimizing the negative log likelihood (NLL) loss:
\begin{align}
\label{eqn:nll}
\gL_{\text{NLL}} (\theta) = -\E_{(\vx,\vy) \sim \data} \left[\sum_{t=1}^T \log \pi_{\theta}(y_t\mid \vy_{1:t-1}, \vx)\right].
\end{align} 
We refer to the model that is trained with the NLL loss as behavioral cloning (BC) \citep{pomerleau1989alvinn} following the RL literature nomenclature.

\section{Reinforced Self-Training (\rest)}
\label{sec:rest}

We present \rest{}, an RLHF algorithm that aligns the language model's outputs with human preferences. Human preferences over sequences are modelled using a learned reward function (see Appendix \ref{app:reward}). In the underlying Markov decision process for conditional language modelling the states are the partial sequences, and the actions are the generated tokens (see Appendix \ref{app:lang_model_rl}).

The \rest{} algorithm decouples the dataset growth and policy improvement of a typical RL pipeline into separate offline stages (Figure \ref{fig:first_figure} and \ref{fig:multi_step_improvement}). We start by training an initial model $\pi_{\theta}(\vy|\vx)$ to map input sequences $\vx$ to output sequences $\vy$ on a given dataset of sequence pairs $\data$ using the NLL loss from Equation~(\ref{eqn:nll}). Next, the \grow{} step creates a new dataset $\data_g$, which augments the initial training dataset with samples from the model:
\[
\data_g = \{ \; (\vx^i,\vy^i)|_{i=1}^{N_g}  \;\;  \mbox{such that}  \;\; \vx^i \sim \data,  \; \vy^i \sim \pi_{\theta}(\vy|\vx^i) \; \} \cup \data. 
\]
Here, the conditioning inputs are resampled from the original dataset $\vx^i \sim \data$, as in self-training, but in situations where one has access to $p(\vx)$ they could sample directly from it, i.e., $\vx^i \sim p(\vx)$. For example, consider a model that generates image from a textual description, in this case, the distribution of text inputs can be sampled from a language model $p(\vx)$.

Subsequently, the \improve{} steps use $\data_g$ to fine-tune the policy $\pi_\theta$. Note that we keep the original dataset in the training mixture to ensure that the policies do not diverge. Below, we describe \grow{} and \improve{} steps in more details.

\newpage
\subsubsection*{\grow{}}
\begin{wrapfigure}[28]{r}{0.38\textwidth}
    \centering
    \vspace{-5mm}
    \includegraphics[width=0.84\linewidth]{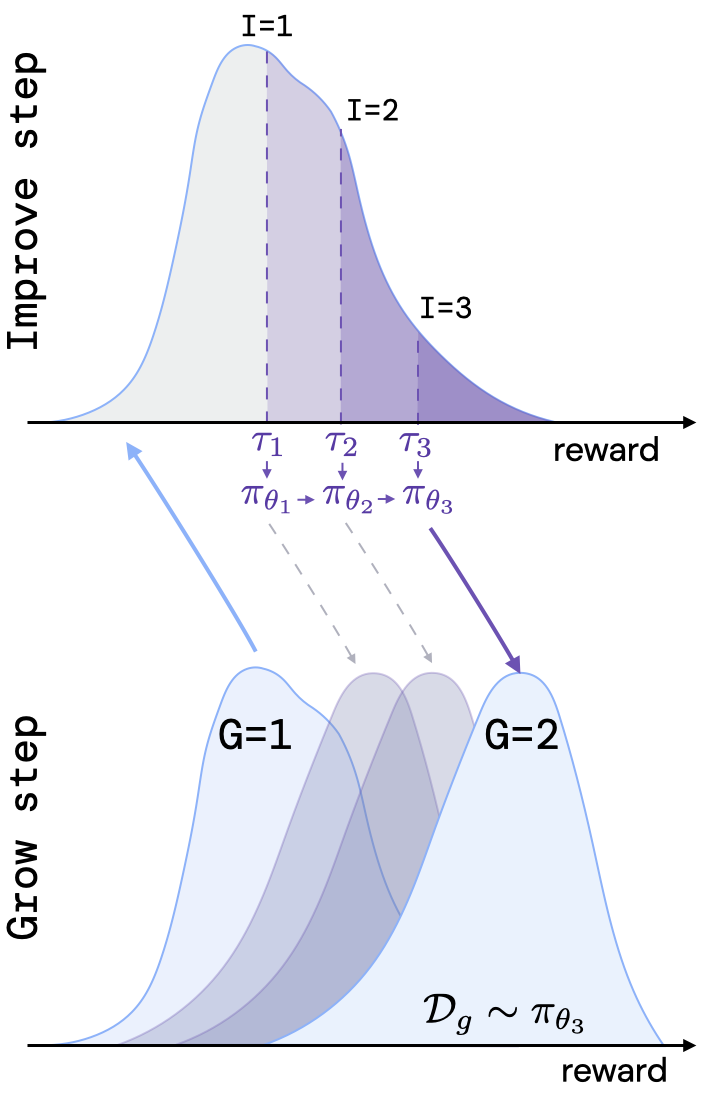}
  \caption{\textbf{\rest{} algorithm}. \textit{Top}: At \improve{} steps \texttt{I=1,I=2,I=3}, the dataset from the initial policy is filtered with thresholds $\tau_1<\tau_2<\tau_3$ and a sequence of policies $\pi_{\theta_1}, \pi_{\theta_2}, \pi_{\theta_3}$ are fine-tuned. \textit{Bottom}: If we were to sample from those policies (grey), the quality of samples would increase. In practice, only the final policy $\pi_{\theta_3}$ is used to generate the next dataset $\data_g$.}
    \label{fig:multi_step_improvement}
\end{wrapfigure}
The \grow{} step corresponds to the acting or data-generation step in RL. We create an augmented dataset of trajectories $\data_{g}$ by sampling many output sequences from the current policy $\pi_{\theta}$, i.e., $\vy \sim \pi_{\theta}(\vy|\vx)$ for $\vx \sim \data$. The new dataset of sequences is then scored with a reward function $R(\vx, \vy)$. The datapoints with the reward above a threshold score are used to update the policy (see next). Once the policy is improved, a new dataset of better quality samples can be created once again (Figure \ref{fig:multi_step_improvement}, bottom). 

\subsubsection*{\improve{}}
At the \improve{} step (exploitation or policy improvement in RL terminology), the goal is to use the new dataset $\data_g$ to fine-tune the policy $\pi_\theta$. We start by defining a filtering function that includes only samples with rewards higher than a certain threshold $\tau$: 
\[
F(\vx, \vy; \tau) = \mathbbm{1}_{R(\vx, \vy) > \tau}. 
\]
Let us note that the threshold based filtering function may result into learning suboptimal behaviors that favors outcomes with high variance in the environments with stochastic dynamics \citep{brandfonbrener2022does}. However, in this work we formulate the language modeling and translation tasks as deterministic RL problems (Appendix \ref{app:lang_model_rl}.)

Next, we finetune the current best policy typically trained with either the supervised learning loss $\gL_{\text{NLL}}$ from~\eqref{eqn:nll} or an offline RL loss $\gL(\vx, \vy; \theta)$ on the filtered data such as V-MPO \citep{song2019v} or offline actor-critic \citep{mathieu2021starcraft}. To sum up, we use the following reward weighted loss $J$:
\begin{align}
\label{eqn:reward_based_loss}
J(\theta) = \E_{(\vx,\vy) \sim \data_{g}} \left[ F(\vx, \vy; \tau) \; \gL(\vx, \vy; \theta)\right].
\end{align}

Standard imitation learning approaches, such as BC (\cite{pomerleau1989alvinn}, \eqref{eqn:nll}) and one-step RL methods like Behavior Value Estimation (BVE) ~\citep{gulcehre2021regularized} perform one-step of \improve{} on the fixed dataset $\data$. In contrast, the basic version of \rest{} additionally includes a \grow{} step that allows the model to gather multiple new output sequences (potential translations) for contexts $\vx$ from the original dataset (source sentences to translate).

When iterating over \improve{} steps, we increase the filtering thresholds: $\tau_1< \dots < \tau_{N-1} < \tau_{N} $ (Figure \ref{fig:multi_step_improvement}). This filtering with the growing threshold results in data subsets of increasing quality but of decreasing size. As LLMs overfit to small datasets quickly, we fine-tune every new policy from the previous policy with a lower learning rate. Consecutive fine-tuning of policies $\{\pi_{\theta_k}\}_{k \geq 1}$ on higher quality data subsets ensures policy improvement with a fixed dataset $\mathcal{D}_g$. If we were to sample from policies $\{\pi_{\theta_k}\}_{k \geq 1}$, the average reward of the generated samples would be increasing (shown in grey in Figure \ref{fig:multi_step_improvement}). As sampling from a policy in the \grow{} step is computationally expensive, after each such step we perform several \improve{} steps. Thus, the cost of a single dataset generation is amortised over multiple \improve{} steps. Algorithm \ref{algo:multi_step_rest} outlines the full \rest{} algorithm with multiple dataset growth and policy improvement steps.

\RestyleAlgo{ruled}

\begin{algorithm}[h]
\KwIn{$\data$: Dataset, $\data_{eval}$: Evaluation dataset, $\gL(\vx, \vy; \theta)$: loss, $R(\vx, \vy)$: reward model, $G$: number of grow steps, $I$: number of improve steps, $N$: number of samples per context}
Train $\pi_\theta$ on $\data$ using loss $\gL$.\\
\For{$g = 1$ to $G$}{
\textcolor{dm-blue-500}{\texttt{// Grow}} \\
Generate dataset $\data_{g}$ by sampling: 
$\data_g = \{ \; (\vx^i,\vy^i)|_{i=1}^{N_g}  \;\;  \mbox{s.t.}  \;\; \vx^i \sim \data,  \; \vy^i \sim \pi_{\theta}(\vy|\vx^i) \; \} \cup \data. $ \\
Annotate $\data_{g}$ with the reward model $R(\vx, \vy)$. \\
\For{$i = 1$ to $I$}{
\textcolor{dm-purple-500}{\texttt{// Improve}} \\
Choose threshold s.t. $\tau_1 >V_{\pi_\theta}$ for $V_{\pi_\theta}=\E_{\data_{g}}[R(\vx,\vy)]$ and $\tau_{i+1} > \tau_i$. \\
\While{reward improves on $\data_{eval}$}{
Optimise $\theta$ on objective: $J(\theta) = \E_{(\vx,\vy) \sim \data_{g}} \left[ F(\vx, \vy; \tau_i) \; \gL(\vx, \vy; \theta) \right]$
}}}
\KwOut{Policy $\pi_{\theta}$}
\caption{\textbf{\rest{} algorithm.} \rest{} is a growing-batch RL algorithm. Given an initial policy of reasonable quality (for example, pre-trained using BC) iteratively applies \grow{} and \improve{} steps to update the policy. Here $F$ is a filtering function, and $\gL$ is an loss function.}
\label{algo:multi_step_rest}
\end{algorithm}

\paragraph{Probabilistic interpretation of the \improve{} step}
Let us consider the particular choice $\gL = \gL_{\text{NLL}}$, with $\theta^{\prime}$ being the parameters of the model from the last \grow{} step, $\lambda$ the proportion of data sampled from this model in $\data_g$ and a single step of growth. The expression for the gradient in this case takes the following form:
\begin{align}
\label{eqn:gradient}
\nabla J(\theta) = -\E_{\vx \sim \data} \left[ \lambda \E_{\vy \sim \pi_{\theta^{\prime}}(\vy|\vx)} \left[  F(\vx, \vy; \tau)  \nabla \log \pi_{\theta}(\vy\mid \vx) \right]  + (1-\lambda) \E_{\vy \sim p(\vy|\vx)} \left[ F(\vx, \vy; \tau) \nabla \log \pi_{\theta}(\vy\mid \vx) \right]\right].
\end{align} 
The first term on the RHS of (\ref{eqn:gradient}) is similar to an online policy gradient term at the beginning of training when $\theta \approx \theta'$ with $F(\vx, \vy; \tau)$ replacing the state-action value function $Q^{\pi}(\vx,\vy)$, when starting in state $\vx$ and taking sequential actions $\vy$, that is generating synthetic data $\vy$ using policy $\pi_\theta$ in our context. For the second term on the RHS of (\ref{eqn:gradient}), we consider the original data $\data$, but we still ensure that it passes the threshold $F(\vx, \vy; \tau)$. Intuitively, people choose $\data$ for training according to some possibly unknown criteria. In this work, we make the criterion  $F(\vx, \vy; \tau)$  explicit. The last term is therefore a form of offline policy gradients which prevents $\pi_{\theta}(\vy\mid\vx)$ to move too far from $p(\vy\mid\vx)$ which could lead to \emph{model collapse} \citep{shumailov2023curse}. Finally, note the similarity of this approach with self-training \citep{clark2003bootstrapping,scudder1965probability,xie2020self} techniques. We provide a population interpretation (i.e., as $N,N_g \rightarrow \infty$) of \rest{} in Appendix \ref{app:Restpopulation}.

In the following section, we explore how the choice of loss, filtering function and threshold, and synthetic data generated by language policy via sampling (exploration data) empirically affect the performance of the resulting policies $\pi_{\theta}$.

\section{Experiments and analysis}
\label{sec:experiments}

We chose machine translation as a testbed for \rest{} as it is an impactful application of conditional language modeling where established reward models are available, for example, Metric X \citep{freitag-etal-2022-results}, BLEURT \citep{sellam-etal-2020-bleurt} and COMET \citep{rei-etal-2020-comet}. We ran experiments on two common benchmarks: IWSLT 2014 \citep{cettolo2014report}, and WMT 2020 \citep{koehn2020findings}, as well as an internal benchmark dataset which we call \textit{Web Domain} (a version of this dataset was previously used by \cite{ghorbani2021scaling}). These datasets contain a set of sentences in the source language and the corresponding human ``reference'' translation. We selected a different language pair for each dataset to test the generality of the results. We kept a separate validation and test sets with unseen source sentences for the evaluation purposes.

We used Metric X in our experiments, a state-of-art reference-free reward model \citep{freitag-etal-2022-results} which, for a given source text and a proposed translation, outputs a numerical score. We report results in terms of average rewards on samples generated by a policy on the validation set \footnote{Performance on the test set follows the same trends (see Appendix \ref{app:reward}). We also experimented with BLEURT and BLEU scores, and \rest{} improved those scores as well. We noticed that online PPO algorithm can learn to exploit the weaknesses and biases of these two metrics quickly, which can cause the model to generate samples that maximize the rewards but deteriorate the quality of the model's output.}. For the details of the datasets and models, we refer to Appendix \ref{app:data_model}. Also, Table \ref{tbl:data_sources} indicates the size of the datasets by reporting the number of samples per source sentence generated at each \grow{} step. 

\paragraph{Nomenclature} We named variants of \rest{} by the loss type, number of \grow{} steps, and number of \improve{} steps, for example $\texttt{GOLD G=1 I=2}$. With this convention, $\texttt{BC G=0 I=0}$ refers to standard supervised learning, which is trained only on the original dataset $\data$ and performs neither \grow{} nor \improve{} steps. When the loss type is not specified, the \BC{} loss is used, i.e., the model is trained with auto-regressive supervised learning with the NLL loss as typical in training language models. In all plots, we colored supervised learning in grey and \rest{} variants in shades of purple.

\paragraph{Baselines} 
We reported the results with several different offline RL method, including Offline Actor Critic (\texttt{OAC}) \citep{mathieu2021starcraft}, Behavior VMPO (\texttt{BVMPO}), Generation by Off-policy Learning from Demonstrations (\texttt{GOLD}) \citep{pang2021}, and \texttt{BC} \citep{pomerleau1989alvinn} \footnote{Details on our baselines are in Appendix \ref{app:baseline_algos} and the experiments with additional losses are in Appendix \ref{sec:negative_offline_rl}}.

\begin{figure}[t!]
    \centering
    \begin{floatrow}
    \includegraphics[scale=.22]{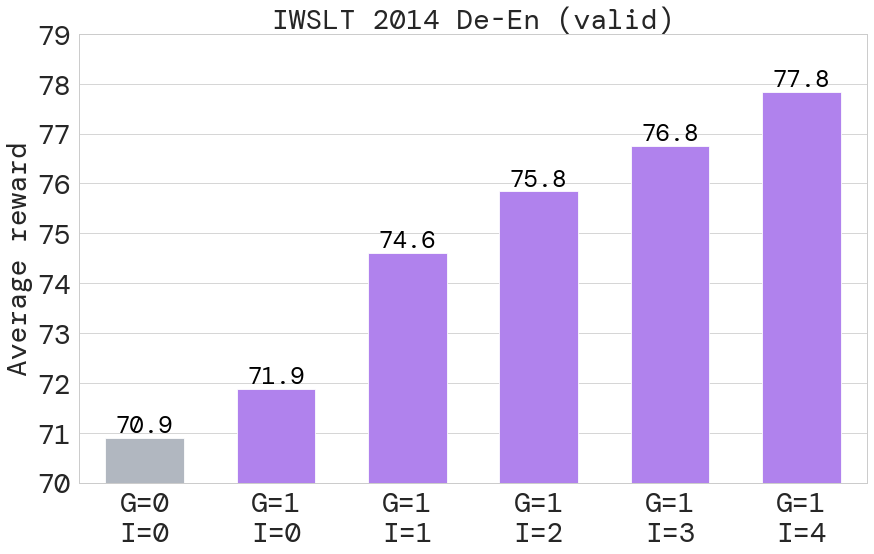}
    \hspace{.2cm}
    \includegraphics[scale=.22]{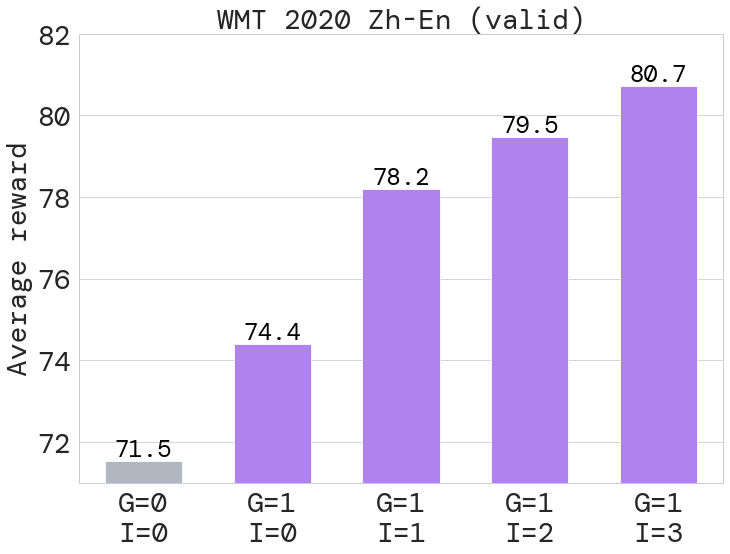}
    \hspace{.2cm}
    \includegraphics[scale=.22]{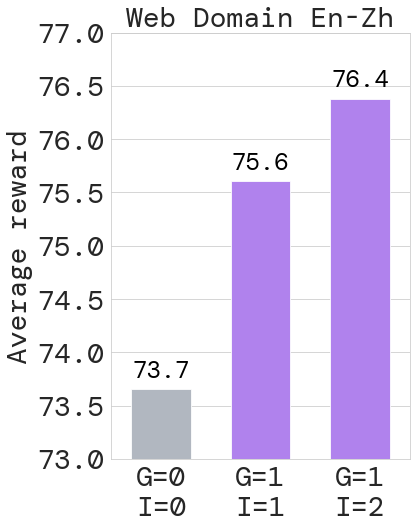}
    \end{floatrow}
    \caption{\textbf{ReST with multiple \improve{} steps.} Average reward model scores on IWSLT 2014 De-En, WMT 2020 Zh-En, and Web Domain En-Zh validation sets. On each dataset, we report results with \BC{} ($G=0,~I=0$) and ReST with a single \grow{} step and several \improve{} steps with an increasing reward threshold. Each \improve{} step increases the reward model score in all three validation datasets. We found the suitable number of \improve{} steps to be a dataset-dependent hyperparameter.
    }
    \label{fig:first_step_iwslt_rest_thresh}
\end{figure}

\paragraph{Do multiple \improve{} steps in \rest{} increase the reward model scores?} 
We evaluated \rest{} on three different datasets by fixing the loss function to \texttt{BC} and increasing the number of \improve{} steps.
The range of rewards for training was normalized between $0$ and $1$ \footnote{Note that the plots show rewards between $0$ and $100$.}. For our experiments, we picked the filtering thresholds $\tau_i$ from a sequence of increasing values $\left[0.0, 0.7, 0.8, 0.9, 0.95, 0.99 \right]$ \footnote{If we used less than 6 steps, we skipped the initial smaller thresholds. Details are given in Appendix \ref{app:data_model}. We ensured that in all our datasets $\tau_1 > 0.7 \ge V_{\pi_\theta}$ by empirically measuring $V_{\pi_\theta}$ over the dataset.}. 
The $\tau_0 = 0.0$ case corresponds to using the full dataset. We did five \improve{} steps on IWSLT 2014, four on WMT-2020, and two on Web Domain. In Figure \ref{fig:first_step_iwslt_rest_thresh} we plotted the average reward of different variants of \rest{}. We see that each subsequent \improve{} step improves the performance of the translation model significantly across all three datasets.

\begin{figure}[t]
    \centering
    \begin{floatrow}
    \includegraphics[scale=.23]{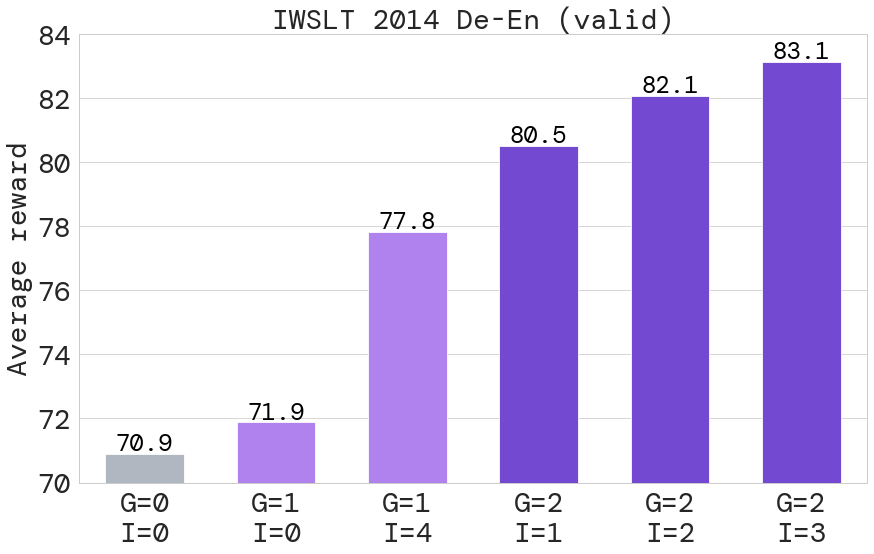}
    \hspace{.2cm}
    \includegraphics[scale=.23]{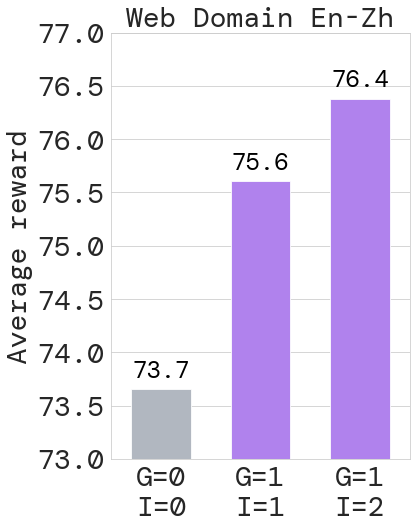}
    \end{floatrow}
    \caption{\textbf{ReST with two \grow{} steps.} The second \grow{} step with subsequent \improve{} steps improves the performance by $5.3$ points on IWSLT 2014 De-En and $0.8$ points on Web Domain En-Zh task over the first \grow{} step.}
    \label{fig:secon_step_rest_thresh}
\end{figure}

\begin{wrapfigure}[17]{ht}{0.44\textwidth}
    \centering
    \vspace{-4.5mm}
    \includegraphics[scale=.22]{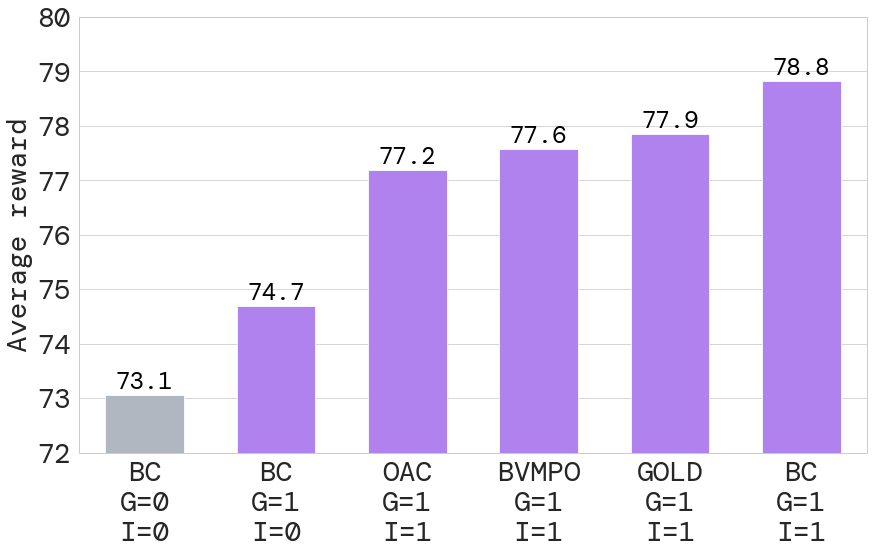}
    \caption{\textbf{WMT 2020 zh-en (test)}: \BC{} (in grey, $G=0~I=0$) and \rest{} trained with different offline RL losses. \rest{} is trained with one \grow{} and \improve{} step except $G=1~I=0$, which is trained on the entire dataset generated after the first \grow{} step without any \improve{} (all in purple). All variants of \rest{} outperform the initial \BC{} baseline, with \BC{} loss resulting in the best performance.}
    \label{fig:baseline_comps}
\end{wrapfigure}

\paragraph{Do additional \grow{} steps improve reward model scores?} We performed a second \grow{} step with successive \improve{} steps to measure the effect of the extra \grow{} step on the performance. In Figure \ref{fig:secon_step_rest_thresh}, a method with an additional \grow{} step achieves further improvement on the IWSLT 2014 and Web Domain datasets. We noticed a $5.3$ point improvement between the end of the first and the second \grow{} step.

\paragraph{Does \rest{} improve over supervised training?} To answer this question, in Figure \ref{fig:baseline_comps} we plotted the average reward achieved by the supervised learning model as well as several variants of \rest{} with different losses and the number of \grow{} and \improve{} steps. Different variants of \rest{} (purple) significantly outperform supervised learning (gray) even after just the first grow step. This observation was consistent across different datasets and language pairs that we tested.

\paragraph{Which loss is the best for a single step of \rest{}?} Figure \ref{fig:baseline_comps} depicts variants of \rest{} with different offline RL losses $\gL(\vx, \vy; \theta)$. We find that \BC{} loss outperforms other loss functions. Note that normally \BC{} algorithm does not depend on the reward, but in \rest{}, the reward is taken into account through the reward filtering stage for $I \geq 1$ (with $\tau_1=0.8$ for WMT 2020.) Results with multiple \grow{} and \improve{} steps are displayed in Figure \ref{fig:secon_step_rest_thresh} (see also Appendix \ref{app:test set results}).

\begin{wrapfigure}[18]{t}{0.4\textwidth}
    \centering
    \includegraphics[scale=0.25]{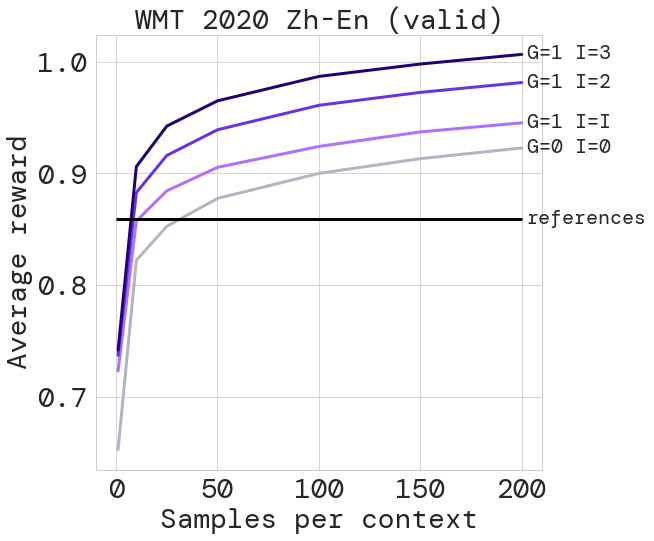}
    \caption{\textbf{Best-of-N sampling at inference time.} All variants of \rest{} benefit as much from Best-of-N sampling as supervised models.}
    \label{fig:wmt_reject}
\end{wrapfigure}

\paragraph{Can \rest{} be improved further with Best-of-N sampling at inference time?} Best-of-N sampling technique at inference time generates $N$ samples which are then ranked by the reward model. Then, the top ranked candidate is selected \citep{gao2022scaling}. We show results with Best-of-N sampling on top of \BC{} ($\texttt{G=0 I=0}$) and \rest{} variants in Figure \ref{fig:wmt_reject}. The performance of \rest{} improves both with $N$ and with the number of \improve{} steps. The best \rest{} variant with $N<10$ matches the performance of the \BC{} model with $N=200$. Even though RL is known to limit the diversity of samples, this experiment shows that \rest{} can still benefit from Best-of-N sampling. After three \improve{} steps with $N=200$, \rest{} achieves the highest possible reward of $1$, outperforming the ``reference'' translations in $\data$.

\paragraph{How does \rest{} compare with Online RL?} 
We compared \rest{} with PPO \citep{schulman2017proximal}, an online RL algorithm widely used for RLHF \citep{ouyang2022instructgpt, glaese2022improving}. For our online RL experiments, we used the setup of \cite{donato2021} where PPO had access to a similar amount of training data as \rest{} with $1$ \grow{} step. The results are summarized in Table \ref{tab:online_rl}. Online RL performs as well as \rest{} with one \grow{} and no \improve{} steps which is equivalent to \BC{} on the $\data_g$ dataset. With the same amount of training data, \rest{} with multiple \improve{} steps achieves significantly higher rewards. Furthermore, we noticed that the BLEU score for the online RL policy on the validation set dropped by nearly $8$ points (BLEU score of \rest{} did not change) which indicates a potential reward hacking behaviour. \rest{}'s ability to improve the reward model score without deteriorating the performance on other metrics suggests that the ``alignment tax'' it pays is lower than for online RL approaches.

\begin{table}[]
    \centering
    \begin{tabular}{c|c|c}
      Algorithm   & Average Reward & Distinct samples \\
      \hline
      \BC{} (G=0, I=0)  & 70.9 & \num{16 000 000} \\
      \rowcolor{lightgray} \rest{} (G=1, I=0) & 71.9 & \num{16000000} \\
      \rest{} (G=1, I=4) & 77.8 & \num{16000000}\\
  \rowcolor{lightgray} \textbf{     \rest{} (G=2, I=3)} & \textbf{83.1} & \num{32000000}\\
      Online RL & 71.6 & \num{24000000} \\
    \end{tabular}
    \caption{\textbf{Online RL for IWSLT 2014:} Online RL performs as well as \rest{} (G=1, I=0) and \rest{} (G=1, I=4) is significantly better.}
    \label{tab:online_rl}
\end{table}

\begin{figure}[h]
    \centering
    \begin{floatrow}
    \includegraphics[scale=.28]{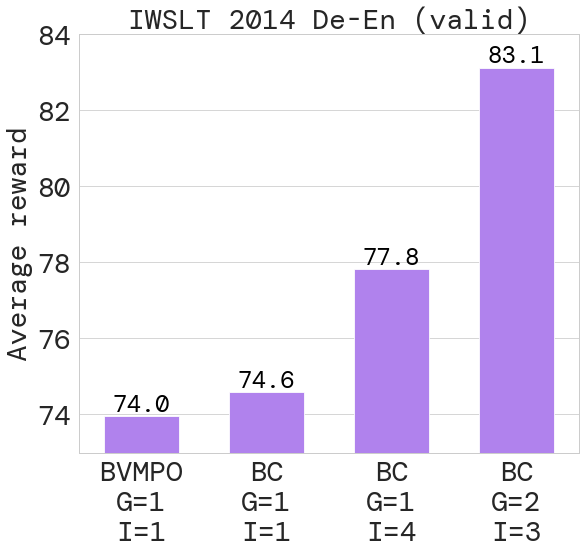}
    \hspace{.2cm}
    \includegraphics[scale=.28]{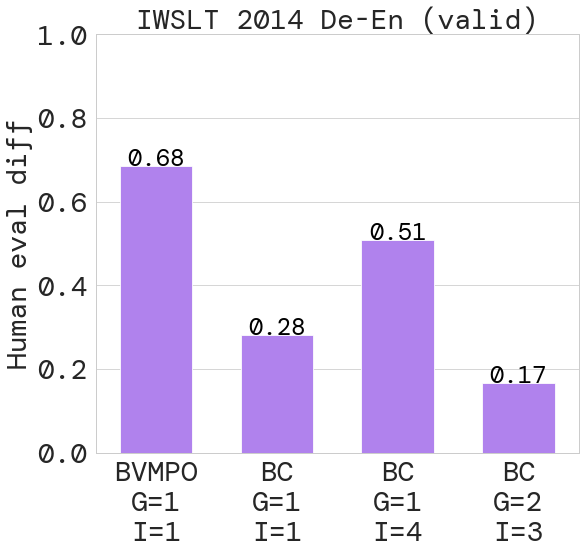}
    \end{floatrow}
    \caption{\textbf{Comparison of performance based on learned reward and on human evaluation.} All \rest{} variants outperform BC in terms of human ratings, but the rankings of the methods based on reward model scores and human scores are different.}
    \label{fig:human_eval}
\end{figure}

\paragraph{Does \rest{} improve human preferences?}
We evaluated the \rest{} models by human raters to investigate if \rest{} can outperform \BC{} in human evaluations as well. We displayed a sentence in the source language and two generated translations: one by \BC{} model ($\texttt{G=0 I=0}$) and one by a \rest{} variant. Human raters scored each translation on a scale from $0$ to $6$, and we measured the difference between the average score of \rest{} method and of \BC{} which we refer as \emph{"Human eval diff"}. In Figure \ref{fig:human_eval} (right), we see that all variants of \rest{} outperform the \BC{} baseline significantly. However, if we compare the human score gains with the gains in the learned reward (Figure \ref{fig:human_eval}, left), the rankings do not match. We hypothesise that the difference is due to the fact that the reward models cannot generalize well on OOD data since the learned reward models are an imperfect proxy of human preferences. In particular, we found that the reward models generalise worse as our policy moves away from the behaviour model which can happen as the number of \grow{} and \improve{} steps increases at which point \rest{} can start overfitting to the reward model. Thus, in our analysis, we focused on evaluating models based on how well they align with a reward signal and we treat reward model generalisation as an independent issue that could be mitigated by, for example, finetuning the reward model between the consecutive \grow{} steps on the human-annotated data from the most recent policy. \rest{} with the B-VMPO loss utilises a learned value function and KL regularisation term to prevent over-fitting and thus attains high human preference scores.

\section{Related works} 

\begin{wrapfigure}[10]{h}{0.5\textwidth}
    \centering
    \vspace{-5mm}
    \includegraphics[scale=.13]{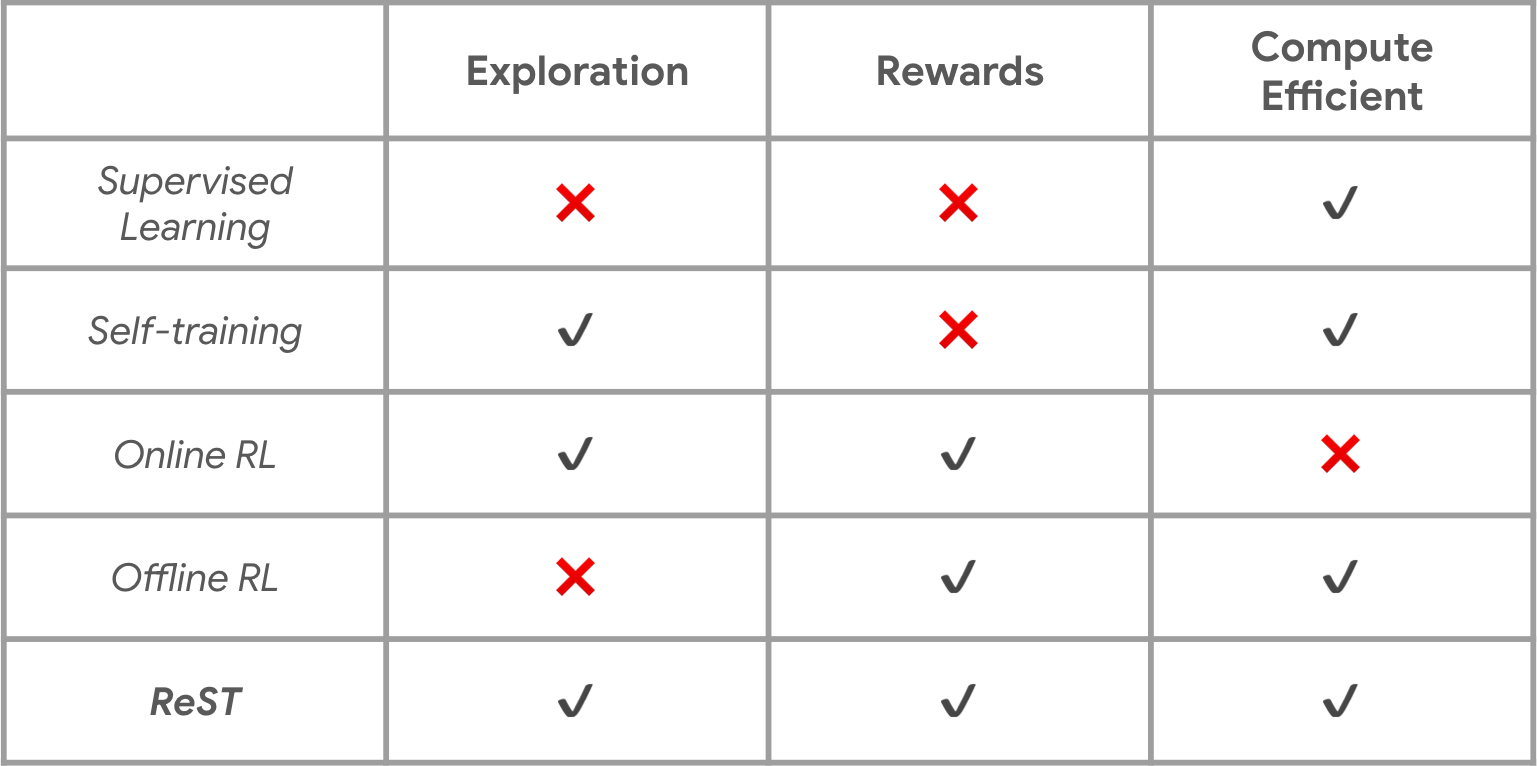}
    \caption{\textbf{\rest{} vs alternatives:} \rest{} is the only approach that can leverage the exploration data and rewards, but is also computationally efficient.}
    \label{fig:framework_comp}
\end{wrapfigure}

There has been large number of works recently on self-improving alignment algorithms for language modelling. In Figure \ref{fig:framework_comp}, we also compare \rest{} against different approaches: supervised learning, self-training, online and offline RL. We conclude from this comparison that \rest{} is the only approach that is compute efficient, but also can leverage exploration data and rewards. Next, we describe some of the particular works related to \rest{}. 

\paragraph{Self-training} Self-training is an established semi-supervised learning approach which utilizes unlabeled data to improve a model \citep{scudder1965probability}. Since its introduction, self-training has been successfully applied to many tasks, including image classification and recognition \citep{xie2020self}, protein folding \citep{jumper2021highly}, as well as several language tasks \citep{yarowsky1995unsupervised,sun2020self,zhang2016exploiting,he2019revisiting}. \cite{he2019revisiting} empirically demonstrated that noisy self-training improves the performance of translation models. The \improve{} step of \rest{} resembles self-training. \rest{}'s main difference from self-training is that the \grow{} step of \rest{} generates synthetic exploration data for training with RL.

\paragraph{Expert Iteration (EI)} \cite{anthony2021expert} proposed an RL framework which is a form of policy iteration approach that makes use of a planning mechanism. EI explicitly decomposes the RL problem into two parts: planning and generalisation. Similar to \rest{}, EI uses the policy to generate data and exploit it to learn a policy with RL. Unlike EI, \rest{} does not require any planning mechanism and it makes use of iterative \improve{} steps that enables it to leverage the gathered data more effectively. 

\paragraph{Reasoning with language models} The EI method inspired several related approaches \citep{zelikman2022star, uesato2022solving}. \cite{zelikman2022star} proposed a technique called STAR that iteratively leverages a small number of rationales to fine-tune the model. Then, they sample rationales with the answers from the model and filter the generated answers by their correctness. \cite{uesato2022solving} proposed a similar method which learns to solve math problems using learned reward models. Recently, \cite{jung2023impossible} proposed a method called ``Impossible Distillation'' similar to ours, which generates datasets for large language models by sampling from a suboptimal model and filters the low-quality examples with a filtering mechanism. This approach corresponds to \rest{} with a single \grow{} and \improve{} steps. In contrast to these methods, \rest{} can be trained with any offline RL losses, with or without planning and various filtering mechanisms. It can also deal with continuous-valued reward scores. Furthermore, \rest{} can perform iterative policy improvement on a fixed dataset with the \improve{} steps.

\paragraph{Iterated Learning (IL)} IL is the process where an agent learns its behavior by being exposed to another agent’s behavior, which itself learned it in the same way \citep{kirby2014iterated}. Recently, this approach was adopted for interactive language learning using deep learning \citep{lu2020countering,lu2020supervised}. IL differs from \rest{} as it operates in a multi-agent setting and does not use RL.

\paragraph{Self Imitation Learning (SIL)} SIL learns a policy for an off-policy actor-critic algorithm where the policy tries to reproduce the good behaviour demonstrated by the agent \citep{oh2018self}. SIL achieves it by filtering out the unsuccessful trajectories from the replay buffer and training the agent only on the high-reward trajectories. In that sense, \rest{} can be considered to be closely related to the SIL-based approaches. The main difference is that \rest{} is agnostic to the underlying RL algorithm used to train the policy and, unlike SIL, does not necessitate a value function to filter out the unsuccessful trajectories. Also, \rest{} is applied to generative AI settings, which do not require interactions with an environment in an online fashion.

\paragraph{Reward ranked Fine-Tuning (RAFT)} Concurrently to our work, \cite{dong2023raft} proposed RAFT. RAFT can be interpreted as a particular case of \rest{} which uses only one \improve{} step for each \grow{} step, and relies on a filtering threshold which is a fixed quantile of the empirical distribution of the rewards of the current samples. %
The authors reported improvements over BC and PPO on a variety of language modeling and image generation tasks. Our experiments with \rest{} showed that multiple \improve{} steps with an increasing filtering threshold for one \grow{} step lead to further performance improvements.

\section{Discussion}

In this paper, we proposed an algorithm called \rest{} which is simple, has minimal hyper-parameters to tune, and is flexible to work with many designs of \grow{} and \improve{} steps. We studied the performance of \rest{} in machine translation as robust and established reward models are available for this task. We experimented with different offline RL losses in the \rest{} loop, but found BC to perform the best for improving the reward model scores. Multiple steps of NLL training with progressively increasing filtering thresholds in the \improve{} step lead to continuous improvements in the model's reward on the holdout set. However, improvements in reward model scores do not necessarily reflect human preferences since the reward model is only a proxy for human preferences. The results indicate that one \grow{} step is the best option when considering human evaluation scores, even though rewards continue to grow with more \grow{} steps. To overcome this limitation, the reward models could be fine-tuned on the subset of \data${_g}$ annotated with human preferences similar to \cite{bai2022training} and \cite{glaese2022improving}, which we leave as future work. Let us note that the risk of overfitting to the reward model increases with the repeated iterations of the \grow{} steps; thus we believe it is essential to address this issue, especially in cases where multiple \grow{} steps are needed to train the model.

As we have seen, simple BC loss still outperforms many offline RL losses in terms of aligning the model with the reward model scores. However, we found that BC can overfit to the reward model so we explain it by the fact that learning value functions in RL is challenging due to sparse rewards, credit assignment problems, sensitivity to hyperparameters, and limited exploration in the \grow{} step. \rest{} could benefit from better RL exploration strategies at \grow{} step, such as MCTS \citep{leblond2021machine}. The ability to exploit the generated data during the \grow{} step could result in a broader exploration of the state-action space and better generalization. Additionally, the determinism of the environment does not allow for large gains over BC for offline RL losses.

To conclude, \rest{} is a general and efficient approach. It can be applied when 1) a robust reward model of human preferences is available and 2) we are able to generate samples from the model at scale. Thus, it can be applied to many tasks within the language domain, such as summarization, turn-based dialogue, and other generative audio and video models. With several avenues for future exploration and applications, we believe that \rest{} is a useful growing batch RL methodology for RLHF.

\paragraph{Acknowledgements}

We would like to thank the members from the machine translation teams at Google and Google DeepMind for their inputs to the project during the brainstorming phases and for setting up the codebase that this project was developed upon \citep{yu2020wmt}. We would like to thank Matt Hoffman, Bobak Shahriari, Taylan Cemgil and Chris Dyer for the discussions about this project. We are grateful for the feedback provided by Bilal Piot for an early draft of this paper. We would also like to thank those responsible for various different frameworks that we used during the project such as the DeepMind JAX ecosystem \citep{deepmind2020jax} and Launchpad \citep{yang2021launchpad}. 

\bibliography{main}
\appendix
\section{Appendix}

\subsection{RLHF for conditional language modeling as MDP}
\label{app:lang_model_rl}
We can formulate conditional language modeling as a sequence to sequence problem. The goal is to map a source sequence $\vx = \left(x_1, x_2, ...x_L\right)$ into a target sequence $\vy = \left(y_1, y_2, ....y_T\right)$, that is to learn a mapping from $\vx$ to $\vy$. Machine translation is a classic example of a sequence to sequence problem \citep{sutskever2014sequence}.

The model can be described as a Markov Decision Process (MDP) \citep{bellman1957}. An MDP, $\gM \defeq (\gS, \gA, \transitions, r, d)$, consists of finite sets of states $\gS$ and actions $\gA$, a transition distribution $\transitions(s'|s,a), s,s'\in\gS, a\in\gA$, a reward function $r:\gS \times\gA \rightarrow \mathbb{R}$, and an initial state distribution $d: \gS \rightarrow [0,1]$. In the offline setting the agent learns from a dataset $\data$ which contains sequences $\left(s_n, a_n, r_{n+1}\right)$. The dataset $\data$ is often assumed to be generated by an unknown {\it behaviour policy} $\mu$ characterised by a distribution over actions conditioned on the state: $\mu(a|s)$. The elements of MDP have the following meaning in conditional language modeling:
\begin{itemize}
    \item \textbf{States} ($s$): The state $s_n$ consists of the input sequence and the partially generated output up to the $n$-th token: $s_n = [\vx, \vy_{1:n-1}]$.
    \item \textbf{Actions} ($a$): The actions are the tokens $a_n = y_n$ to be generated by policy ($\pi$).
    \item \textbf{Rewards} (r): Rewards can be given or learned. In our experiments we train a deep neural network on human preferences to assigns a score to the full generated sequence. Reward is produced after end-of-sequence (EOS) token, and it is zero at all other steps.
   \item \textbf{Transition operator} (T): The transition operator defines the dynamics of an environment: $\transitions(s^\prime|a,s) := \transitions \left(y_n | s_n = [\vx, \vy_{1:n-1}]\right)$. In our setup, it is a deterministic operator that concatenates newly produced token (action) to the current sequence (state).
\end{itemize}

This RLHF formulation of conditional language modeling can be also seen as a contextual bandit problem with a very large action space. %

\subsection{Negative results with offline RL}
\label{sec:negative_offline_rl}
In addition to the experiments from Section \ref{sec:experiments}, we also run experiments with Q-learning with BVE type of one-step RL approaches and reward conditioned approaches, such as decision transformers \citep{chen2021decision}. However, we could not obtain notable improvements with them over the supervised baseline. Q-function-based methods for learning a policy and value function performed worse than supervised learning when used as a policy even after training from initialization from a supervised checkpoint. This matches previous observations by \cite{mathieu2021starcraft} who showed that offline Q-learning has difficulty in large action spaces  (vocabulary size of \num{32000} tokens) and state-value based methods are preferable. Furthermore, as  \cite{brandfonbrener2022does} showed, return-conditioned methods tend to learn sub-optimal policies on tasks with sparse continuous rewards.

\subsection{Data and Model details}
\label{app:data_model}
In all our experiments, the base policy architecture is a minor modification of a standard Transformer architecture \citep{vaswani2017attention}. We use a vocabulary of \num{32000} tokens and a max sequence length of \num{128} at decoding time. We use the \texttt{sentencepiece} tool \citep{kudo2018subword} for building the tokenizers.

\paragraph{\grow{} step}
During the \grow{} step, we sampled from the latest checkpoint of the policy with tempered softmax using temperature $0.8$ following the procedure proposed by \cite{li2022competition} to generate the dataset. Morevero, in our analysis, we found that temperature $0.8$ often covers a broad range of rewards in the dataset.

\paragraph{Thresholds in \improve{} step}
If we set the threshold $\tau$ to a value such that $\tau  \geq  V^{\pi_0}$, and learn a policy $\pi^{1}$ on data filtered with this threshold, the updated average value $V^{\pi_1}$ of this policy satisfies $V^{\pi_1} > V^{\pi_0}$. Then, iterative threshold ensures policy improvement in the \improve{} step. In our experiments, we do several improvement steps until we reach the maximum threshold. At each step, we trained the model for \num{500000} SGD steps and pick the checkpoint with the best reward model score.

\paragraph{IWSLT 2014 De-En}
We use train, validation and test sets from IWSLT 2014 De-En dataset \citep{cettolo2014report} which includes source sentences in German with human translations (references) in English. In each \grow{} step, we generate \num{100} candidate translation for each source sentence in the training set, effectively yielding us $|\data_g|=$\num{16000000}. For this dataset, we use a tiny version of the Transformer encoder-decoder \citep{vaswani2017attention} architecture with the feedforward MLP layers of size \num{512}, feedforward dimension of \num{1024}, \num{4} attention heads and \num{6} encoder and decoder layers.

\paragraph{WMT 2020 Zh-En}
We use the source-reference pairs in Chinese and English from the work of \cite{koehn2020findings} for our training, validation and test sets. Exact details on the datasets and preprocessing can be found in \cite{yu2020wmt}. In each \grow{} step, we generate \num{25} candidates for each source sentence in the training set, effectively yielding us $|\data_g|=$ \num{890000000}. We choose to generate \num{25} candidates per source as the WMT dataset is significantly ($\sim$ 100 times) larger than the IWSLT dataset.  We use an architecture that mimics the Transformer-base encoder-decoder \citep{vaswani2017attention} architecture with model dimension \num{512}, feedforward dimension of \num{2048}, \num{8} attention heads and \num{6} encoder and decoder layers.

\paragraph{Web Domain En-Zh}
Finally, we use the in-house dataset for English to Chinese translation with custom training, fine-tuning and test sets. This training corpus is our biggest dataset \footnote{For computational reasons, we run \rest{} on the fine-tuning corpus with a model warm-started from a supervised model trained on the entire massive training corpus.}, so we use a modified version of Transformer-big encoder-decoder \citep{vaswani2017attention} architecture with model dimension \num{1024}, feedforward dimension of \num{8192}, \num{16} attention heads and \num{6} encoder and decoder layers.

In Table \ref{tbl:data_sources}, we list all the datasets with their sizes. In all the experiments, unless stated otherwise, we report the average reward scores on the validation set.
 \begin{table}[ht!]
 \caption{Details of the datasets and their sizes used in \rest{} experiments.}
 \begin{tabular}{@{}lrrrl@{}}
 \toprule
 Dataset    & $|\data_g|$ & \# Eval samples & \# Candidates\\& & & per source \\ \midrule
 IWSLT 2014 de-en & \num{16000000}       &       \num{7466}          & \num{100}                      \\
 WMT 2020 zh-en  & \num{890000000}      &       \num{2000}          & \num{25}                       \\
 Web Domain Finetuning en-zh  & \num{3000000}        & \num{6000}                 & \num{1000}                     \\ \bottomrule
 \end{tabular}
 \label{tbl:data_sources}
 \end{table}

\subsection{Reward model}
\label{app:reward}
We used learned reward models that assign a score to the whole translation. We considered two types of reward models for translation \citep{freitag-etal-2022-results}: i) \textit{Reference-free} reward models estimate of how good a translation is based on the source and candidate sentences only, ii) \textit{Reference-based} reward models additionally use the reference human translation to decide how good a candidate translation is. The reference-free reward models are more flexible since they do not require reference translations. Moreover, by their design, reference-based models assign higher scores for sentences similar to the reference, and this could happen even when the reference is erroneous or incomplete. Finally, reference-free models open possibilities to evaluate and discover candidate translations that supersede the quality of references. Thus, in our experiments, we chose to work with reference-free reward models. On the other hand, since the reference-free reward models are not grounded in a human reference translation, it is more vulnerable to the distribution shifts in the candidates generated by the model and reward hacking. In practice, we pre-computed and stored $R(\vx, \vy)$ for the generated $\data_{g}$.

During the development of the model, we relied on the methodology of ``unit tests'' for reward models where we tested them in various hand-crafted settings, such as on various permutations and repetitions of sentences. The unit-tests ensured the high quality of rewards generated for \rest{}. Even though we used the most robust available reference-free reward model, we found that it was still not perfect, and sometimes showed signs of delusions. For example, the reward of a translation occasionally increased when we repeated the sentence, independent of the initial quality of the translation.

\subsection{Alignment between human evaluation scores and the reward model scores}
\label{app:human_evals}

\begin{figure}[htbp!]
    \centering
    \includegraphics[width=0.8\linewidth]{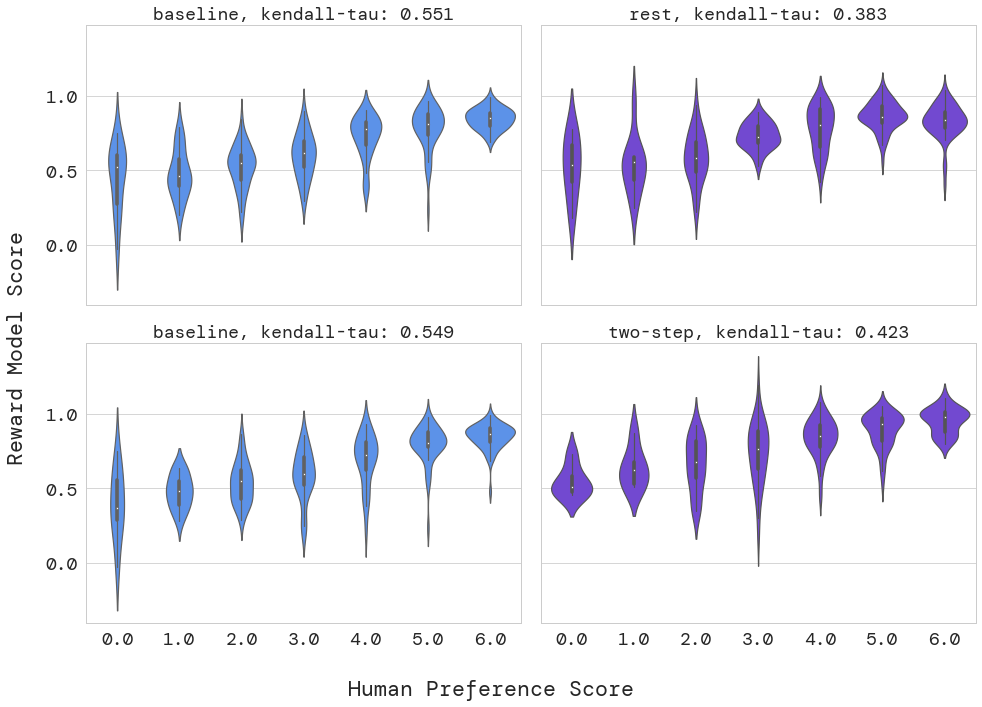}
    \caption{\textbf{[IWSLT 2014 De-En]:} distribution of human preference and reward model scores for \rest{} (BC, $\texttt{I=4, G=1}$) in side by side evaluation with supervised model. The human preference scores less than or equal to \num{3} have significantly higher variance in rewards compared to the human preference scores above \num{3}. The plots on the left-hand side are for the samples generated from a supervised baseline and the right-hand side are for the samples generated with \rest{}.}
    \label{fig:iwslt_rest_violin_human_evals}
\end{figure}

\begin{figure}[htbp!]
    \centering
    \includegraphics[width=0.8\linewidth]{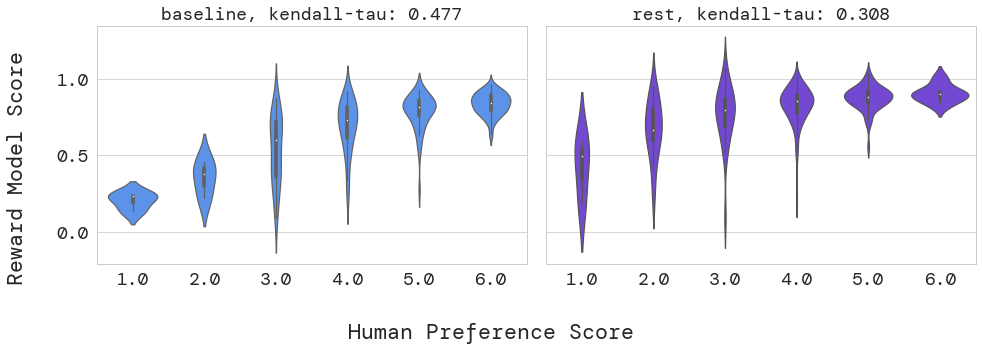}
    \caption{\textbf{[WMT 2020 Zh-En]:} distribution of human preference and reward model scores for \rest{} (BC, $\texttt{I=4, G=1}$) in side by side evaluation with supervised model. The human preference scores lower than or equal to \num{3} has a variance in terms of reward model scores. The reward model has less certainty on the scores of the candidates having lower scores.}
    \label{fig:wmt_rest_violin_human_evals}
\end{figure}

In Figure \ref{fig:iwslt_rest_violin_human_evals} and \ref{fig:wmt_rest_violin_human_evals}, we showed the distribution of human preferences and reward model scores for \rest{} with \BC{} and $\texttt{G=1 I=4}$. Here, we found out that our reward model has a very high variance on samples with low human preference score from \rest{} (\BC{}, $\texttt{G=1 I=4}$) and our supervised learning baseline (\BC{}, $\texttt{G=0 I=0}$). Our hypothesis for this is that the training dataset for the reward model was dominated by translations of high quality \footnote{The data came from WMT 2021 metrics shared task dataset \citep{freitag2021results}, which reviews the human evaluation of good translation models}. Thus the model overfitted to the samples with high scores, and it has high variance in its prediction over the samples with lesser quality. This gap can be addressed with the incremental retraining of the reward model.

\subsection{Results on test sets}
\label{app:test set results}
\begin{figure}[ht!]
    \centering
    \begin{floatrow}
    \includegraphics[scale=.24]{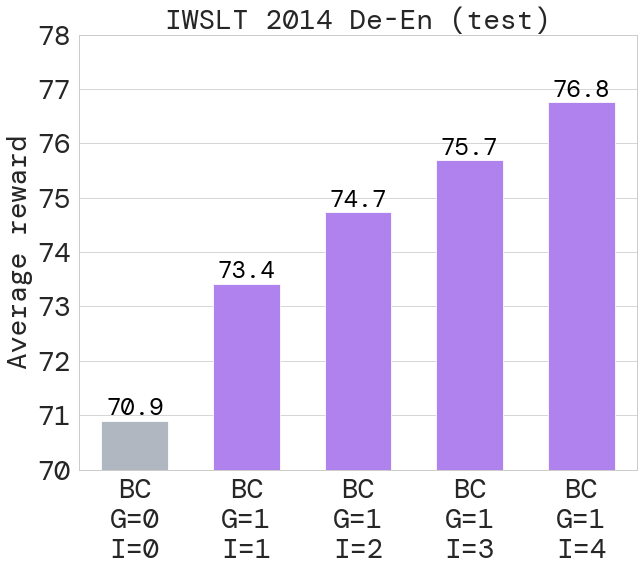}
    \hspace{.3cm}
    \includegraphics[scale=.24]{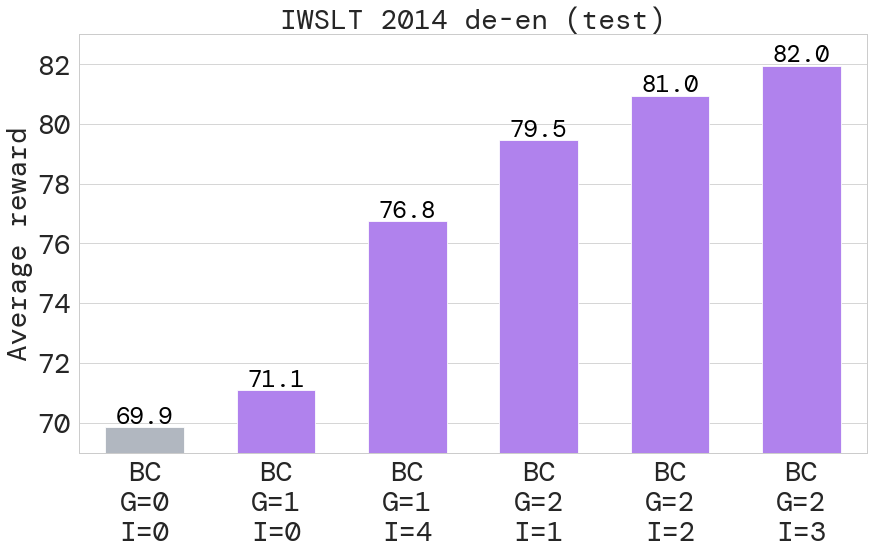}
    \end{floatrow}
    \caption{\textbf{[IWSLT 2014 De-En]: ReST results on test set.} All variations of \rest{} (purple) outperform supervised learning baseline (grey). As we increase the number of \grow{} and \improve{} steps, the reward of the model on the test set continues to increase. 
    }
    \label{fig:improve_steps_rest_test}
\end{figure}

\begin{figure}[ht!]
    \centering
    \begin{floatrow}
    \includegraphics[scale=.28]{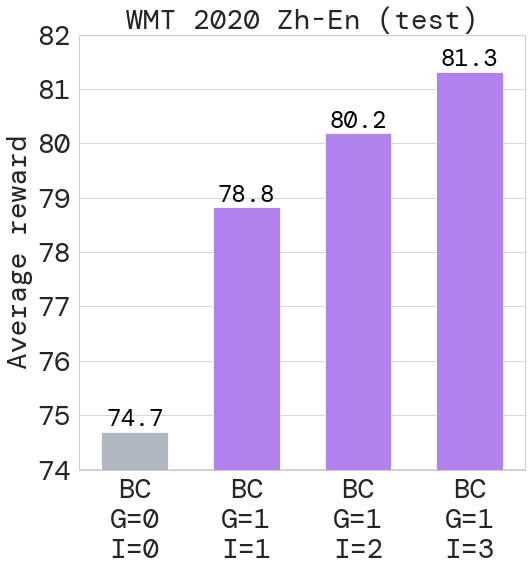}
    \hspace{.3cm}
    \includegraphics[scale=.28]{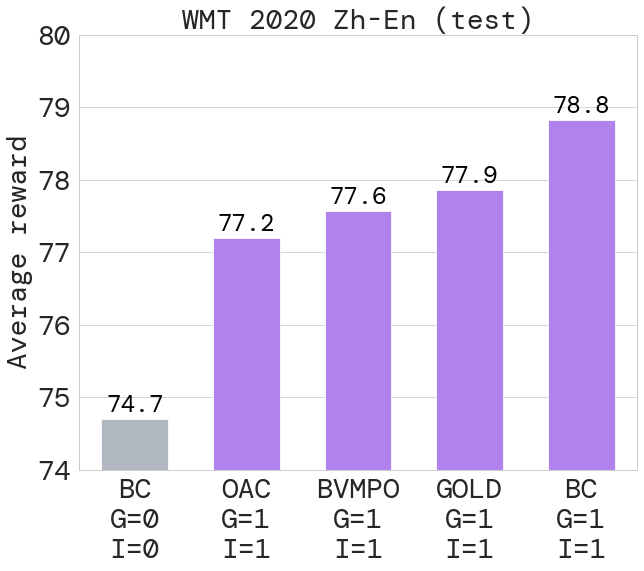}
    \end{floatrow}
    \caption{\textbf{[WMT 2020 Zh-En]: ReST results on test set.} We see that $\rest$ clearly outperforms the baselines (right) and the reward of the model on the test set increases with the number of \improve{} steps.  
    }
    \label{fig:wmt_rest_test}
\end{figure}

Figures \ref{fig:improve_steps_rest_test} and \ref{fig:wmt_rest_test} present the results on the test sets in IWSLT 2014 De-En and WMT 2020 Zh-En datasets. Similar to the validations set, $\rest$ outperforms other value-based offline RL baselines. Also, increasing the number of both \grow{} and \improve{} steps helps to further increase the rewards. 

\subsection{Additional experiments}
This section reports the results of several ablations in \rest{} which explain some of the design choices that we made. 

\begin{figure}[!ht]
    \RawFloats
    \centering
    \begin{minipage}{0.45\linewidth}
    \centering
    \includegraphics[width=0.96\linewidth,height=0.25\textheight]{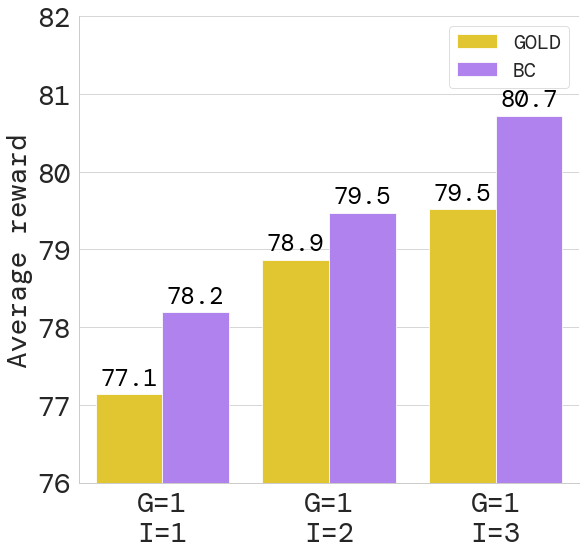}
    \caption{\textbf{[WMT 2020 Zh-En]:} GOLD loss vs BC loss in multi-step improvement. The performance of all methods improves with more steps, but BC loss always performs better than other offline RL losses that we tried.}
    \label{fig:wmt_gold_vs_bc}
    \end{minipage}%
    \hspace{4mm}
    \begin{minipage}{0.45\linewidth}
    \centering
    \includegraphics[width=0.96\linewidth,height=0.27\textheight]{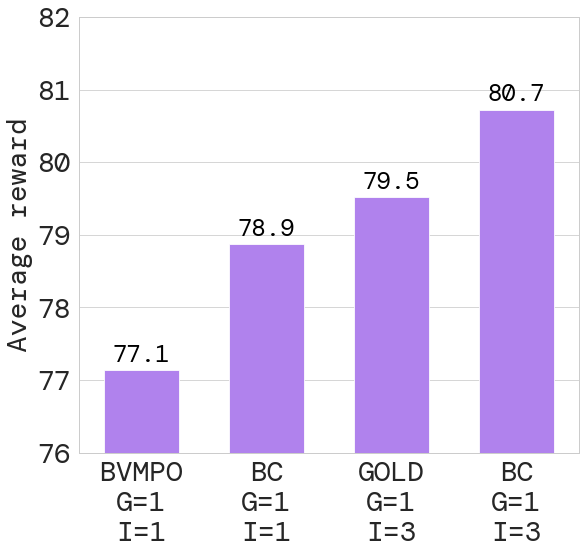}
    \caption{\textbf{[IWSLT 2014 De-En]:} BVMPO, BC, GOLD losses used in \rest{}. BC loss with three \improve{} steps yields the best results.}
    \label{fig:iwslt_baseline_comparisons}
    \end{minipage}
\end{figure}

\paragraph{GOLD loss compared to BC loss in \improve{} step} 
In Figure \ref{fig:wmt_gold_vs_bc}, we compare GOLD against BC losses with the increasing number of \rest{} \improve{} steps on WMT 2020 Zh-En dataset. The performance of both BC and GOLD improves with the number of \improve{} steps, but BC performs better than GOLD in every setting.

\paragraph{Loss comparison on IWSLT 2014 De-En}
In Figure \ref{fig:iwslt_baseline_comparisons}, we compare BVMPO, BC and GOLD losses in \improve{} step on IWSLT 2014. The results are consistent with WMT dataset: in short, \rest{} with BC loss and multiple \improve{} steps outperforms other approaches.

\begin{figure}[!t]
    \centering
    \begin{floatrow}
    \includegraphics[width=0.4\linewidth]{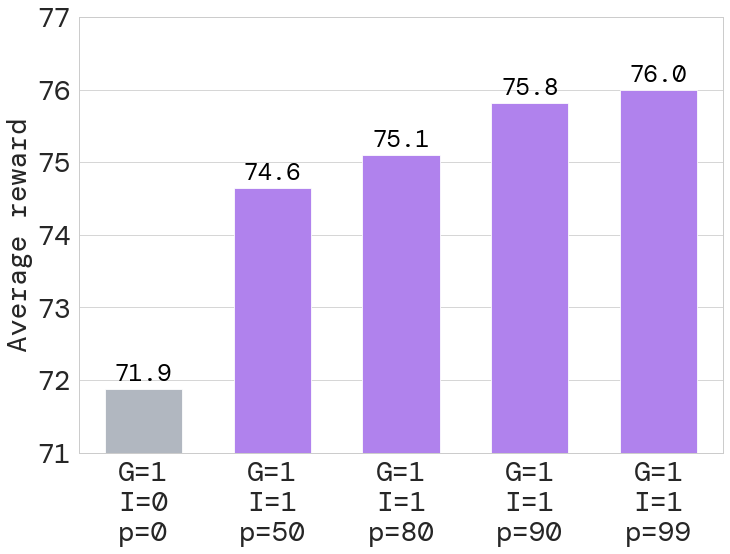}
    \hspace{.3cm}
    \includegraphics[width=0.4\linewidth]{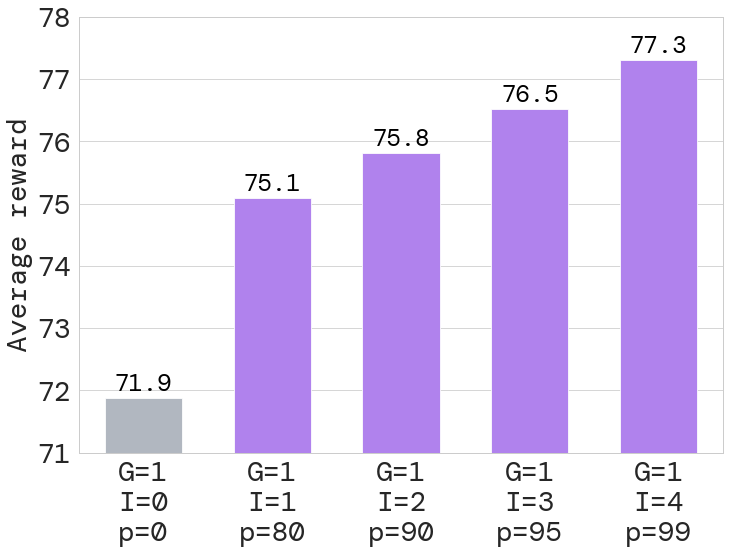}
    \end{floatrow}
    \caption{\textbf{[IWSLT 2014 De-En] percentiles}: Filtering data based on percentiles of reward scores for each source candidate. \rest{} performs better with increasing percentile and the performance saturates after $p=90$.}
    \label{fig:iwslt_percentile_comparisons}
\end{figure}

\paragraph{Selecting threshold based on percentiles of reward model scores}
Using a single threshold for all the source-candidate pairs may lead to a scenario with no training data for certain (harder) source sentences or contexts. Alternatively, we can filter the candidate sentences based on the percentile of the scores given their source. This way of filtering based on percentiles effectively results in changing the thresholds for each source sentence. The results are presented in Figure \ref{fig:iwslt_percentile_comparisons}. As the filtering percentile increases, the performance increases, but it starts saturating, and the gains become smaller after $p=90$. The results are generally similar to those with global threshold-based filtering.

\paragraph{Selecting threshold based on linear interpolation between the mean and max scores of candidates}

\begin{figure}[t!]
    \centering
    \begin{floatrow}
    \includegraphics[width=0.35\linewidth]{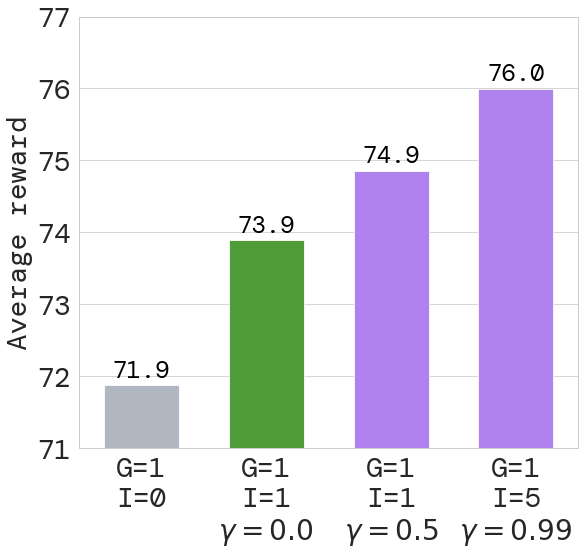}
    \hspace{.3cm}
    \includegraphics[width=0.52\linewidth]{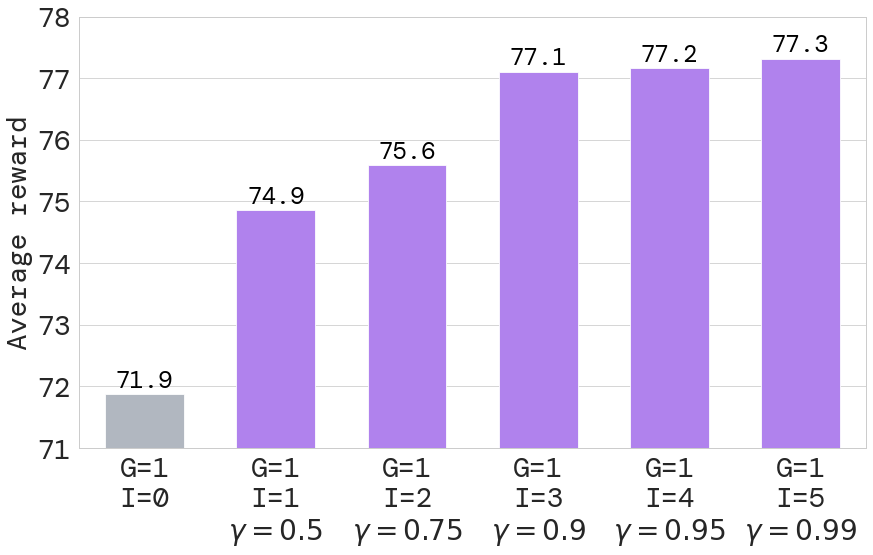}
    \end{floatrow}
    \caption{\textbf{[IWSLT 2014 De-En] interpolation experiments:} Threshold in \improve{} step is chosen for each candidate sentence as $\gamma V^{\max}(source) + (1-\gamma) V^{\text{mean}}(source)$. On the left, experiments with $\texttt{I=1}$ provide improvements and increasing $\gamma$, which also helps substantially. On the right, we present the results with $\gamma$ starting with $0.5$ instead of $0.5$. As it can be seen from this figure, the number of \improve{} steps and how the thresholds are selected greatly influence the final performance of the model. }
    \label{fig:iwslt_interpolation}
\end{figure}

An alternative way of specifying filters based on a source sentence is to compute the filter based on a max reward score ($V^{\max}(source)$) and the mean candidate score ($V^{\text{mean}}(source)$) for a give source sentence. Then, we select the threshold for each source sentence as $\tau_{\gamma}(source) = \gamma V^{\max}(source) + (1-\gamma) V^{\text{mean}}(source)$. At each \improve{} step, we increase the parameter $\gamma$. We report the results with respect to different $\gamma$ values in Figure \ref{fig:iwslt_interpolation}. In a nutshell, computing the threshold by interpolating the max and mean scores for a given candidate gives results similar to the percentile-based way of computing the thresholds per source. Also we can see that the schedule of thresholds and the number of \improve{} steps have a large impact on the final performance of the model.

\subsection{Baseline Algorithms}
\label{app:baseline_algos}

This section provides details on the baseline losses used in Figure~\ref{fig:baseline_comps}.

\subsubsection{BVMPO}
The BVMPO approach is similar to DIME proposed by \cite{abdolmaleki2021multi}. The main difference is that we use a state-value function instead of Q funtion with v-trace \citep{espeholt2018impala}, similarly to V-MPO \citep{song2019v}. We train separate neural networks for policy and value function. The policy is pre-trained with BC and the value network is pre-trained with BVE to estimate the behavior value $V^{\beta}$. BVMPO uses the behavioral cloning policy $\pi^{\beta}(a|s)$ as a behaviour prior:
\begin{equation}
    \gL_{BVMPO} = \gL_{VMPO} + \lambda \textup{KL}(\pi^\beta | \pi).
\end{equation}
As opposed to DIME, BVMPO starts with a fixed coefficient $\lambda=1$ and anneals it to $1e-5$ during the training. We found that starting with a large $\lambda$ and annealing it stabilizes the training without any noticeable drop in performance. Similarly, annealing the behavior regularization co-efficient was also shown to be effective in offline Q-learning \citep{agarwal2022beyond}.

\subsubsection{GOLD}
The GOLD loss was introduced by \cite{pang2021}. When using it in \improve{} step, we start with a BC policy. In all GOLD experiments, we used $k=5$ steps to unroll the MLE model into the future for reward computations. Increasing $k$ did not improve the performance.

\subsubsection{Offline Actor Critic (OAC)}
Offline actor critic is an actor critic approach proposed by \cite{mathieu2021starcraft} where the state-value function is trained to estimate the behavior value in the dataset \citep{gulcehre2021regularized}. Unlike \cite{mathieu2021starcraft} we did not use the V-trace as it did not improve over the vanilla advantage function in our initial experiments.

\subsubsection{Value Functions}

We trained state-value functions on IWSLT 2014 De-En and WMT 2020 Zh-En datasets after the initial grow step. The value functions estimate the behavior value of the policy on the training dataset. The value functions predict the sum of discounted rewards for each word until the end of the sentence. We used a transformer-based encoder-decoder architecture for the value functions. The encoder gets the source sentence as the input, and the decoder predicts Monte Carlo discounted returns conditioned on the source and history of generated sequences of tokens in the target.

On the training sets of IWSLT and WMT, our value functions achieved Kendall-tau correlation of $92.4\%$ and Spearman correlation above $99\%$ with respect to the reward model scores in the dataset. As we do RL only offline, the value functions do not need to generalize beyond the training dataset.

\subsection{\rest{}: Population interpretation}
\label{app:Restpopulation}
Here, we provide a population interpretation of the \rest{} algorithm. In this setup, the initial BC training step can be thought of as identifying $\theta$ by minimizing
\begin{align*}
    \textup{KL}(p(\vx,\vy) ||\pi_{\theta}(\vx,\vy))&=\sum_{\vx,\vy} p(\vx,\vy) \log \frac{p(\vx,\vy)}{\pi_{\theta}(\vx,\vy)}\\
    &=\sum_{\vx,\vy} p(\vx,\vy) \log \frac{p(\vy|\vx)}{\pi_{\theta}(\vy|\vx)}\\
    &=\mathbb{E}_{\vx \sim p(\vx)}[ \textup{KL}(p(\vy|\vx) ||\pi_{\theta}(\vy |\vx))],
\end{align*}
where $\textup{KL}$ is the Kullback--Leibler discrepancy, $p(\vx,\vy)=p(\vx)p(\vy|\vx)$ is the (training) data distribution and $\pi_{\theta}(\vx,\vy)=p(\vx)\pi_{\theta}(\vy|\vx)$. Let us call the minimizer $\theta_0$. 

Assume now that one has the parameter $\theta_k$ at the end of \improve{} steps of the $k${th} \grow{} step. Then at the next $(k+1)${th} \grow{} step, we consider \improve{} steps using a sequence of increasing thresholds $(\tau_i)_{i=1}^{I}$ which are used to filter the available data distribution (mixture of original data distribution and synthetic data distribution) to obtain a new distribution 
\begin{equation*}
    \pi_{\theta_k}^{\tau_i}(\vx,\vy) \propto \{(1-\lambda) p(\vx,\vy)+\lambda \pi_{\theta_k}(\vx,\vy)\}\mathbb{I}(R(\vx,\vy)> \tau_i).
\end{equation*}
We then obtain the next parameter $\theta$ by minimizing the KL 
\begin{equation*}
      \textup{KL}(\pi^{\tau_i}_{\theta_k}(\vx,\vy)||\pi_\theta(\vx,\vy))
\end{equation*}
and keep increasing the threshold until the reward $\mathbb{E}_{\pi_\theta(\vx,\vy)}[R(\vx,\vy)]$ stops increasing. This yields $\theta_{k+1}$.
\end{document}